\documentclass[11pt,a4paper]{article}
\usepackage[hyperref]{acl2019}
\usepackage{times}
\usepackage{latexsym}
\usepackage{amsmath}

\usepackage{epsfig}
\usepackage{graphicx}
\usepackage{amsmath}
\usepackage{amssymb}
\usepackage{float}
\usepackage{multirow}
\usepackage{color}
\usepackage{graphics}
\usepackage{comment}
\usepackage{url}
\usepackage{textcomp}
\usepackage{mathtools}

\usepackage{url}

\aclfinalcopy 

\title{Learning to Answer by Learning to Ask: \\Getting the Best of GPT-2 and BERT Worlds}

\author{{Tassilo Klein$^1$, Moin Nabi$^1$} \\
$^1$SAP Machine Learning Research, Berlin, Germany\\
{\tt $\{$tassilo.klein, m.nabi$\}$@sap.com}}

\date{}
 
\begin{document}

\maketitle

\begin{abstract}
Automatic question generation aims at the generation of questions from a context, with the corresponding answers being sub-spans of the given passage. Whereas, most of the methods mostly rely on heuristic rules to generate questions, more recently also neural network approaches have been proposed. In this work,
we propose a variant of the self-attention Transformer network architectures model to generate meaningful and
diverse questions. To this end, we propose an easy to use model consisting of the conjunction of the Transformer decoder GPT-2~\cite{radford2019language} model with Transformer encoder BERT~\cite{DBLP:journals/corr/abs-1810-04805} for the downstream task for question answering. The model is trained in an end-to-end fashion, where the language model is trained to produce a question-answer-aware input representation that facilitates to generate an answer focused question. Our result of neural question generation
from text on the SQuAD 1.1 dataset~\cite{rajpurkar-etal-2016-squad} suggests that our method can produce semantically correct and diverse questions. Additionally, we assessed the performance of our proposed method for the downstream task of question answering. The analysis shows that our proposed generation \& answering collaboration framework relatively improves both tasks and is particularly powerful in the semi-supervised setup. The results further suggest a robust and comparably lean pipeline 
facilitating question generation in the small-data regime.
\end{abstract}

\begin{figure*}[!t]
\centering
\includegraphics[width=.95\textwidth]{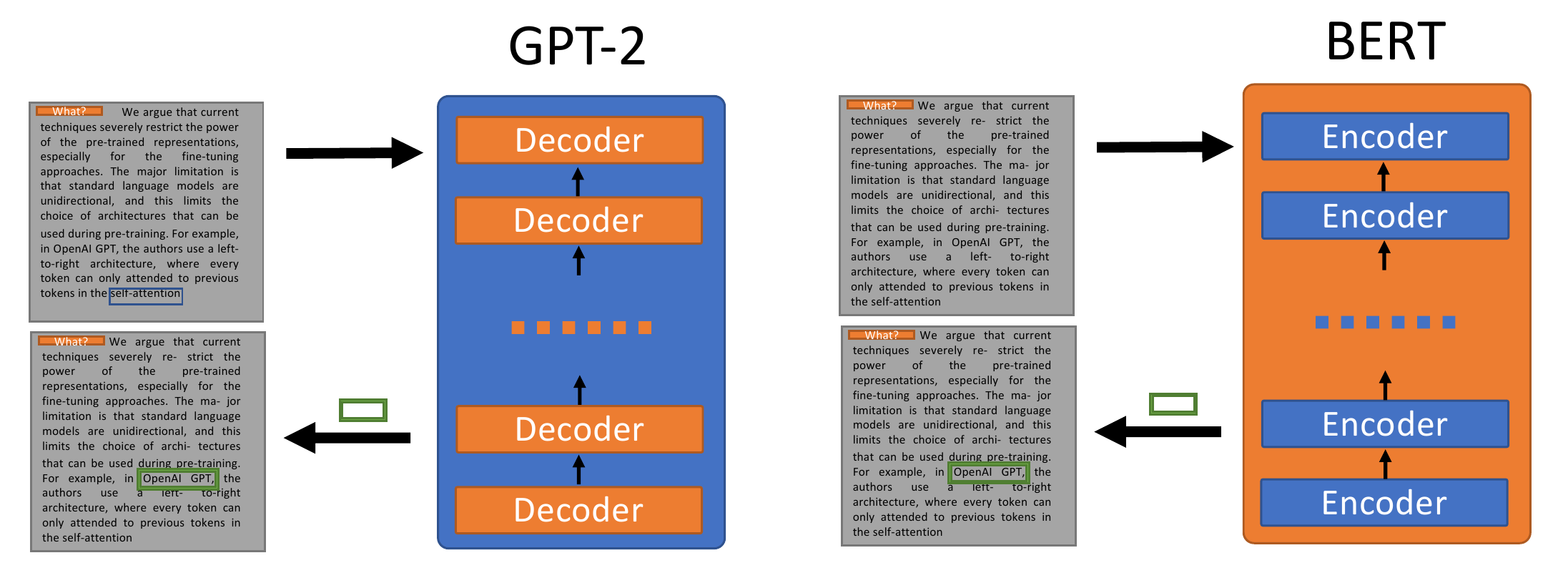}
\captionof{figure}{\textbf{Illustration of the pre-training sketch:} Each network, i.e. GPT-2 and BERT, is individually trained to answer questions using a QA head assigning probabilities to each token to be beginning and/or end of the answer span. The small blue box corresponds to an annotated answer. The small orange square denotes the question, whereas the green box indicates the answer span annotation returned by the models. (Best viewed in color)}
\label{fig:fig-1}
\end{figure*}

\section{Introduction}

Recently, natural language processing (NLP) has enjoyed unprecedented progress largely due to developments in deep learning. In this regard considerable attention in the NLP community is devoted to topics related to automatic question answering (QA). In comparison to that, the inverse task - automatic question generation - has received significantly less attention. Although, question generation (QG) \cite{du2017learning,serban2016generating,pan2019recent,kim2019improving} enjoys a bit of niche existence, it has a plethora of potential applications such as improving the training of QA systems, and help in the creation of material in the educational domain \cite{chen2018learningq}. Automatic creation of questions as well as making them person-specific, can alleviate educator from this tedious task, improving the education experience of both teachers and students.

In this paper, we consider collaborative learning of QA and QG. The key idea of this work is that as question answering and generation are naturally related tasks, so leveraging their connection should be mutually beneficial in  terms of performance as well as reducing the amount of labeled data, e.g. for training a QA system. 
The proposed solution builds upon two recent variants of self-attention Transformer network architectures~\cite{vaswani2017attention}, namely GPT-2~\cite{radford2019language} and BERT~\cite{DBLP:journals/corr/abs-1810-04805}. Essentially, the Transformer architecture consists of two main building blocks, stacked encoders and decoders, which are connected in a cascaded fashion. The encoder is further divided into two components: self-attention layer and a feed-forward neural network. Self-attention allows for attending to specific words encoding and therefore establishing a focus context w.r.t. each word. The decoder has an additional encoder-decoder layer that is switched between the self-attention and the feed-forward network. This allows the decoder to attend to specific parts of the input sequence. 
Compared to the original Transformer architecture, GPT-2 discards the encoder blocks reducing it to a decoder stack. It provides traditional language model functionality, allowing to predict the next word in a sequence given the history. Consequently,  it is naturally applicable in general \emph{generative} tasks. However, since it is not optimized for \emph{question generation} purpose, there is no guarantee that the generated questions are valid and answerable. 
In contrast the latter model BERT is a \emph{masked} language model. It allows only for predicting a masked out word conditioned on both its left and right context, notably establishing word embeddings in a context-specific and bi-directional manner.  What is more, BERT is trained for discriminative QA with applying a specific regression head. Specifically, it predicts the answer text span in the given paragraph for a given question. However, even beyond QA BERT has shown extreme versatility in terms of applicability for numerous downstream tasks.
In comparison to the conventional Transformer network and in contrast to GPT-2, it discards the decoder blocks, reducing it to a pure encoder stack.

This work relates to recent studies which attempt to improve the performance of a discriminative QA model with generative QG models \cite{lewis2018generative,wang2017irgan,duan-etal-2017-question,dong-etal-2017-learning,yang-etal-2017-semi,harrison2018neural,tang2018learning}. These works regard QA as the primary task and use auxiliary tasks, such as question generation and question paraphrasing, to improve the primary task. Whereas this is one part of our goal, our other goal is to improve the QG model with the QA system and to further improve both tasks in a loop. Another key difference compared to these methods is that all these methods require very similar models for the generation and answering, whereas our method is built on top of two different Transformer architectures for answering and generation, coupling the best of both worlds. Specifically, a Transformer network model is proposed which consisting of the conjunction of the Transformer decoder GPT-2 model with Transformer encoder BERT.\\
In a similar work,~\cite{tang2017question} consider learning question generation and question answering as a dual task, but learning them jointly using the entire dataset. Specifically, they employ a triplet-like loss by sampling negative and positive pairs of questions and answers, injecting the duality into the optimization procedure via a regularization term. However, in contrast to our proposed approach, semi-supervised learning is not considered. Manual question generation is a laborious and tedious task. Therefore, employing QG in a semi-supervised setup is very desirable.  The most related work \cite{yang-etal-2017-semi} proposes to use a GRU-based encoder/decoder architecture to generate questions based on the unlabeled text. The generated questions are then combined with the human questions through a domain adaptation pipeline for training QA models.
This approach employs a heuristic on computing possible answers, whereas the proposed approach requires weak labels in terms of answer spans, although in theory could also be operated on heuristic generated annotations.

Existing studies \cite{nema2018towards,zhang2019bertscore} have shown that the task of question generation often exhibits linguistic variance that is semantically admissible; this renders it inappropriate to judge a generated question solely by matching against a ground truth sentence. Therefore, as another contribution of this paper, we propose to assess the quality of QG systems using the performance of a QA network trained on generated questions by QG as surrogate measure. 

The proposed approach is evaluated on the SQuAD dataset~\cite{rajpurkar-etal-2016-squad} for both ``question generation'' and ``question generation'' tasks. In both tasks we show improvement as a result of our proposed generation \& answering collaboration framework. This study opens up avenues for exploiting inexpensive QG solutions similar to ours to achieve performance gain in QA task.

The contributions are two-fold: 
\textbf{First}, we leverage question generation by tying together GPT-2 and BERT in end-to-end trainable fashion facilitating semi-supervised learning. 
\textbf{Second}, we propose to use QA as a surrogate measure for assessing question generation quality.




\begin{figure*}[!t]
\centering
\includegraphics[width=.95\textwidth]{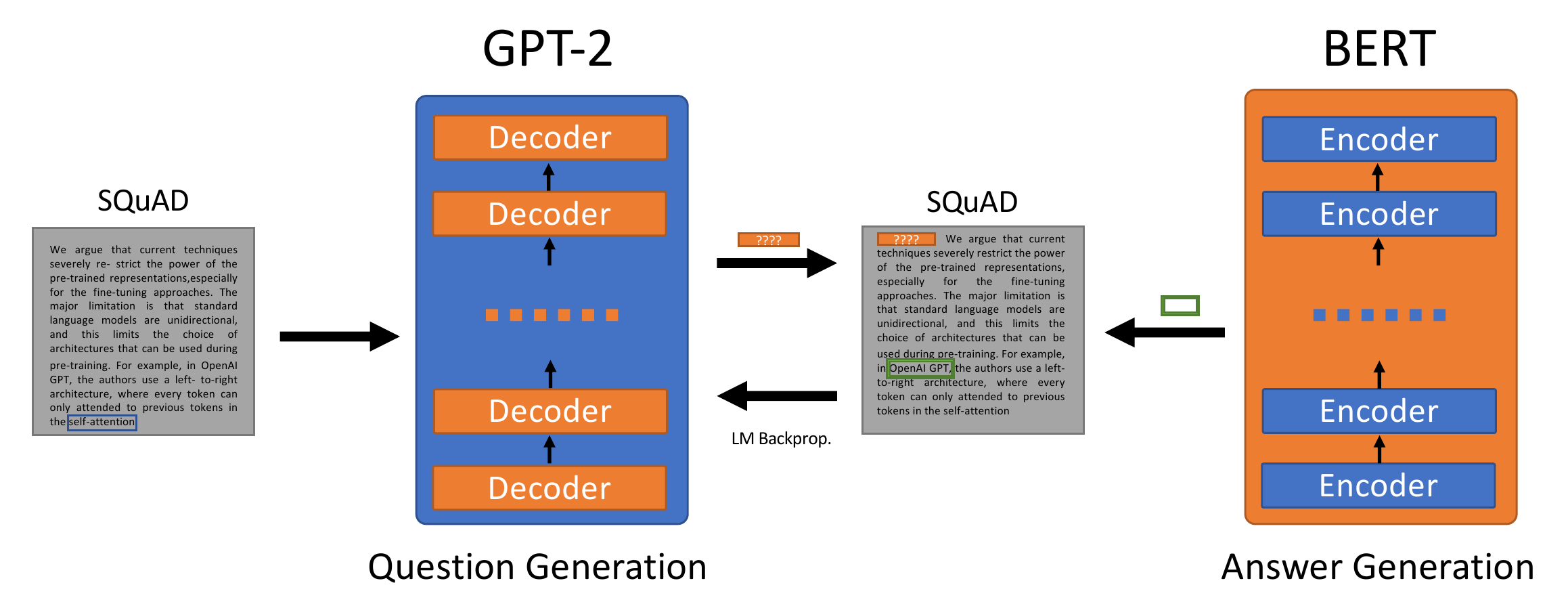}
\captionof{figure}{\textbf{Overview of the fine-tuning of the approach:} Given a SQuAD context and an annotated answer (blue box), a question is generated using GPT-2. The generated answer is denoted with the orange box of question marks. The SQuAD context endowed with the generated question is given to the pre-trained BERT network. BERT then generates an answer span, denoted with green box. If BERT is unable to provide the correct answer, the language model's loss is backpropagated to GPT-2 w.r.t. the annotated context. (Best viewed in color)}
\label{fig:fig-1}
\end{figure*}

\section{Method}
Question answering and question generation are intrinsically linked. Therefore, it is natural to combine these aspects together to improve the desired task. Hence, the core of the proposed method consists of learning question generation network by making use of the feedback of a question answering network. 

Here we propose to employ GPT-2's language model for question generation and BERT for question answering. 
We first briefly discuss the intrinsics of GPT-2 and BERT. This is followed by elaborating on how we adapt GPT-2's language model, 
for question generation. Subsequently, we explain the details on how to 
merge the process of question generation \& question answering in a collaborative framework through mixing GPT-2 and BERT.

\subsection{Background}
In this section we briefly review GPT-2~\cite{radford2019language} and BERT~\cite{DBLP:journals/corr/abs-1810-04805} models. Both are variants of self-attention ``Transformer'' network architectures~\cite{vaswani2017attention}. The former provides a traditional language model, allowing to predict the next word in a sequence given the history. In contrast to that, the latter is a masked language model. It allows for predicting a masked out word, conditioned on both, its left and right context. Another key technical innovation in BERT is its bidirectional training instantiating a context specific word embedding. Context-specific embeddings are in stark contrast to static word embeddings such as word2vec~\cite{word2vec:2013}. In this regard, BERT can easily be fine-tuned for a plethora of different downstream tasks. This can largely be attributed to the self-attention mechanism in the Transformer that allows BERT facilitates generic applicability. Another interesting aspect of BERT is that it does not have an explicit notion of word order beyond marking each word with its absolute-position embedding.

Wang and Cho~\cite{DBLP:journals/corr/abs-1902-04094} also showed BERT is a combination of a Markov random field language model with pseudo log-likelihood training. As a consequence, similar to a traditional language model, this formulation automatically allows for Gibbs sampling sequences. 
Technically, GPT-2 and BERT are opposite slices of the Transformer stack. That is, GPT-2 incorporates the Decoder stack of the Transformer architecture, whereas BERT consists of the Transformer Encoder stack. 

\subsection{Question Generation and Answering}
For question generation with GPT-2, we follow the standard strategy for text generation as proposed in the original paper~\cite{radford2019language}. Given the natural sequential ordering of the language model, the joint probability of a sequence $\mathbf{s}=\left(s_1,...,s_n\right)$ can be factorized into a product of conditional probabilities
\begin{equation}
p\left(\mathbf{s}\right)=\prod_i^n p\left(s_n\mid s_1,...,s_{n-1}\right),
\label{eq:seq}
\end{equation}
largely following~\cite{JelMer80,Bengio:2003:NPL:944919.944966}. This in turn allows for efficient sampling strategies such as sequential top-k~\cite{fan-etal-2018-hierarchical,radford2019language} (here, k=1). At each sampling step, the model computes the word probability over the entire vocabulary for being the next word. This is followed by  randomly sampling from the k most-likely candidates. Sampling is discontinued when a maximum sequence length is reached or a special terminal symbol is produced, e.g. ``?'' for questions.

\begin{table*}[htp]
\centering
\begin{tabular}{lllllll}
\hline
Method & BLEU-1 & BLEU-2 & BLEU-3 & BLEU-4 & ROGUE-L \\
\hline\hline
QA-QG-Dual \cite{tang2017question} & - & - & - & 5.03 & -  \\
LM-init \cite{radford2019language} & 24.85 & 17.85 & 11.06 & 6.85 & 33.56  \\
Our Proposed Method & \textbf{31.46} & \textbf{19.50} & \textbf{12.41} & \textbf{7.84} & \textbf{34.51}  \\
\hline
\end{tabular}
\caption{Quality of generated questions on SQuAD 1.1, with BLEU and ROGUE metric. BLEU score for the approach of~\cite{tang2017question} taken from their paper, which is comparable but computed on a different split.}
\label{tab:results-quantitative-metrics}
\end{table*}

However, in order to be tailored to the specifics of questions a number of extensions have to be made. Specifically, as GPT-2 is trained for general text generation a fine-tuning stage has to be included, which allows for the conditional generation of questions given an annotated possible answer. To this end, the model's tags are augmented by special tokens delimiting answers during training. Thus during the training phase, a question context $c$ is provided together with $l$ answer-question tuples $\left(a_i,q_i\right)$, whereby $l$ varies from context to context, and $a_i$, $q_i$ denoting the groundtruth answer question, respectively. Furthermore, we denote the length for the groundtruth answer as $m_i = \vert q_i \vert$. 
Then during the optimization, we maximize the likelihood $Q$ over all contexts and their respective tuple sets denoted as \begin{equation}
X = \bigcup_{1,..,u}\left\{ (q_1, a_1),...,(q_k, a_k)\right\},
\label{eq:qa_set}
\end{equation}
where $u$ denotes the context cardinality. 
Thus, factorizing over all contexts, we yield

\begin{equation}
Q=\prod_{k}^{u}\prod_{k}^{l_k}\prod_{i}^{m_{k,j}}p\left(s_{m_{k,j}}\mid s_1,...,s_{{k,m_j}-1}; c_k, a_{k,j}\right),
\end{equation}

where in contrast to Eq.~\ref{eq:seq} conditioning is extended by a context $c_k$ and a specific answer in the context $a_{k,j}$.

The fine-tuning step yields a model that allows for basic QG (i.e., \textsc{LM-init} in the experiment). However, in order to boost the performance with increased diversity in generation output, we have a subsequent downstream optimization step that ties together question generation with a QA module. Details about the collaborative downstream optimization are explained in the next section.

\subsection{Collaborative Generation \& Answering}
The models trained to this point consist of a rudimentary GPT-2 network question generation and a BERT network for question answering, respectively. 
The next step entails fine-tuning of both models in tandem in an end-to-end fashion. The underlying idea is to exploit the duality of the tasks in order to increase the diversity of the answer generation, specifically capitalizing on the QA power of BERT. Thereby, the QA module is employed statically for the task of question answering, whereas GPT-2's task is to generate questions which are improved over time. That is, backpropagation is only performed w.r.t. weights of the QG module, namely the GPT-2 language model weights, whereas the weights of QA remain unchanged. Technically, it is easily possible also to perform backpropagation w.r.t. weights of the QA module, however, this increases the risk of drift and unstable behaviour (loss oscillations) during optimization such that regularization becomes non-trivial. Furthermore, experiments indicate that a short finetuning step is sufficient for the QA network to serve as feedback mechanism.  

Specifically, given a context with annotated question, a question is generated using GPT-2 - identically as discussed in the previous section. The difference lies in the forthcoming steps. Then the context endowed with the newly generated question (without answer annotation) is given to the pre-trained QA network. The BERT QA network then in turn generates an answer span, which is compared with the groundtruth. A question that cannot be answered by the QA system, i.e. yields an incorrect answer span, gives rise to sub-optimal wording or semantic mismatch.
Therefore, we backpropagate the loss incurred by these samples w.r.t. GPT-2 language model. Effectively, this leads to a division the tuple set $X$ ( see Eq.~\ref{eq:qa_set}) in two different parts during optimization. Namely, we obtain
\begin{equation}
X = X_{-a}\cup X_{a} \quad\textbf{s.t.}\quad X_{-a}\cap X_{a} = \emptyset.    
\end{equation} 
One set $X_{-a}$ contains contexts and answers that cannot be answered, and the other set $X_{a}$ contains context-answer pairs that are answerable. These sets represent a performance snapshot of the system at current iteration. Thus, during each round of the optimization we try to continuously shrink the first the cardinality of $X_{-a}$, i.e. minimizing $\vert X_{-a}\vert$ and thus minimizing the number of questions that are answered incorrectly. At the same time, in order to avoid catastrophic forgetting, we keep probing for the previously correctly answered questions (known as replay mechanism in continual learning~\cite{shin2017continual}, continuously sampling from $X_{a}$, trying to maximize $\vert X_{a}\vert$. In case a previously answered question cannot be answered anymore, it moves back to the set of unanswerable question, such that at any time the following holds: $X_{-a}\cap X_{a} = \emptyset$. See Fig. \ref{fig:fig-1} for the illustration of the optimization pipeline.

\begin{table*}[htp]
\centering
\begin{tabular}{l|l|lll}
\hline
 Labeling rate & Method &  Dev F1 &  Test F1 &  Test EM    \\
\hline\hline
0.1 & Gen + GAN~\cite{pmlr-v37-ganin15} & 0.4897 & 0.4373 & 0.2885 \\
0.1 & Gen + dual~\cite{NIPS2016_6469} & 0.5036 & 0.4555 & 0.3005 \\
0.1 & Gen + domain~\cite{yang-etal-2017-semi} &  0.5234 & 0.4703 & 0.3145 \\
0.1 & Gen + domain + adv~\cite{yang-etal-2017-semi} &  0.5313 & 0.4802 & 0.3218 \\
0.1 & Our Proposed Method & \textbf{0.6931} & \textbf{0.6391} & \textbf{0.4741} \\
\hline
0.2 & Gen + GAN~\cite{pmlr-v37-ganin15} & 0.5525 & 0.5037 & 0.3470 \\
0.2 & Gen + dual~\cite{NIPS2016_6469}  & 0.5720 & 0.5192 & 0.3612 \\
0.2 & Gen + domain~\cite{yang-etal-2017-semi} &  0.5749 & 0.5216 & 0.3658 \\
0.2 & Gen + domain + adv~\cite{yang-etal-2017-semi} &  0.5867 & 0.5394 & 0.3781 \\
0.2 & Our Proposed Method & \textbf{07614} & \textbf{0.7053} & \textbf{0.5476} \\
\hline
0.5 & Gen + GAN~\cite{pmlr-v37-ganin15} & 0.6110 & 0.5590 & 0.4044 \\
0.5 & Gen + dual~\cite{NIPS2016_6469}  & 0.6368 & 0.5746 & 0.4163 \\
0.5 & Gen + domain~\cite{yang-etal-2017-semi} &  0.6378 & 0.5826 & 0.4261 \\
0.5 & Gen + domain + adv~\cite{yang-etal-2017-semi} &  0.6375 & 0.5831 & 0.4267 \\
0.5 & Our Proposed Method & \textbf{0.8185} & \textbf{0.7564} & \textbf{0.6056} \\
\hline
0.9 & Gen + GAN~\cite{pmlr-v37-ganin15} & 0.6396 & 0.5874 & 0.4317 \\
0.9 & Gen + dual~\cite{NIPS2016_6469}  & 0.6511 & 0.5892 & 0.4340 \\
0.9 & Gen + domain~\cite{yang-etal-2017-semi} &  0.6611 & 0.6102 & 0.4573 \\
0.9 & Gen + domain + adv~\cite{yang-etal-2017-semi} &  0.6585 & 0.6043 & 0.4497 \\
0.9 & Our Proposed Method & \textbf{0.8409} & \textbf{0.7755} & \textbf{0.6282} \\
\hline
\end{tabular}
\caption{Performance with various labeling rates and methods with unlabeled dataset comprising 50k samples. ``Dev'' denotes the development
set, and ``Test'' denotes the test set, with exact measure (EM) and F1 metric. Results from all approaches apart from the proposed one are taken from~\cite{yang-etal-2017-semi}.}
\label{tab:results-squad_split-1}
\end{table*}

\section{Experiments \& Results}
In order to assess the performance a number of experiments were conducted. At first a qualitative analysis is performed in terms of language generation metrics such as BLEU and ROGUE. Due to the deficiency of these scores as will be discussed, QA is employed as an indirect surrogate measure. Next, the behaviour of QG in a semi-supervised setup is analysed, simulating the feasibility in small-data regime problems. Finally, we perform an ablation study in terms of analyzing the importance of QA component used during training. 

Experiments are performed on the Stanford Question Answering Dataset (SQuAD) v1.1 dataset~\cite{rajpurkar-etal-2016-squad}. It consists of a collection of more than 100k question/answer pairs w.r.t. paragraphs from Wikipedia articles that were acquired by crowdsourcing. 
We employ a data split which divides the training data set equally in two parts. The first part \textbf{(SP2)} is used for supervised pre-training of the QG and QA models. The second half \textbf{(SP1)} is used for evaluation purposes. 

Initialization of the model entails pre-training of both BERT and GPT-2. Whereas BERT is fine-tuned for the task of question answering, GPT-2 is fine-tune for question generation. Both is conducted for 2 epochs.


\subsection{Quality of Generated Questions:}
For question generation, we evaluated the quality of generated questions by comparing it with the groundtruth questions existing in the dataset using the standard language generation metrics: BLUE and ROGUE (Tab. \ref{tab:results-quantitative-metrics}). 

Figure \ref{fig:qual} shows some qualitative results on the question generation. As can be seen, generated sentences have high diversity and differ significantly from groundtruth. The last example shows one case of failure of our method, due to the very fine-grained nature of the question. As can be seen, questions generated by the proposed approach have good quality. Therefore the proposed approach is applicable in the low-data regime, compensating absence of annotations in large corpora. Questions generated also have higher quality than using the vanilla GPT-2 language model, suggesting that learning from BERT in the feedback loop provides further language cues, which may be attributed to the strength of the context-specific embeddings of BERT that allows for establishing complex relationships in sentences as well as rich semantic representation that can be exploited by QA.

\begin{table}[htp]
\centering
\begin{tabular}{l|ll}
\hline
 Method & EM &  F1   \\
\hline\hline
Supervised (Upper-bound) & 79.60 & 87.30 \\
\hline
LM-init \cite{radford2019language} &  67.51 & 77.15 \\
Our Method (\emph{GPT-2})&  70.61 & 79.73   \\
Our Method (\emph{BERT}) & \textbf{75.37} & \textbf{84.42}    \\
\hline
\end{tabular}
\caption{Question answering performance on SQuAD 1.1 (SP1), with exact measure (EM) and F1 metric. Our Method (\emph{BERT}) denotes the proposed approach using GPT-2 for question generation as well as BERT as question answering. Our Method (\emph{GPT-2}) denotes the approach employing GPT-2 for question generation as well as the modified GPT-2 QA module as discussed in the section dealing with the ablation study of the QA component.}
\label{tab:results-squad-generative}
\end{table}

\begin{table}[htp]
\centering
\begin{tabular}{l|ll}
\hline
 Method & EM &  F1   \\
\hline\hline
Supervised (Upper-bound) & 80.80 & 88.50 \\
\hline
LM-init \cite{radford2019language} &  67.51 & 77.15 \\
Our Method  & \textbf{78.47} & \textbf{86.41}    \\
\hline
\end{tabular}
\caption{Question answering performance on SQuAD 1.1 (all), with exact measure (EM) and F1. \emph{Our Method} denotes to the BERT QA model trained on entire training set of SQuAD, but half the training data is fully supervised (SP2), but the other half (SP1) is generated by our proposed method, as discussed in the section dealing with quantitative evaluation using surrogate measure.}
\label{tab:results-squad_full}
\end{table}

\subsection{Quantitative Analysis using Surrogate Measure}



\begin{figure*}[!t]
\centering
\includegraphics[width=1\textwidth]{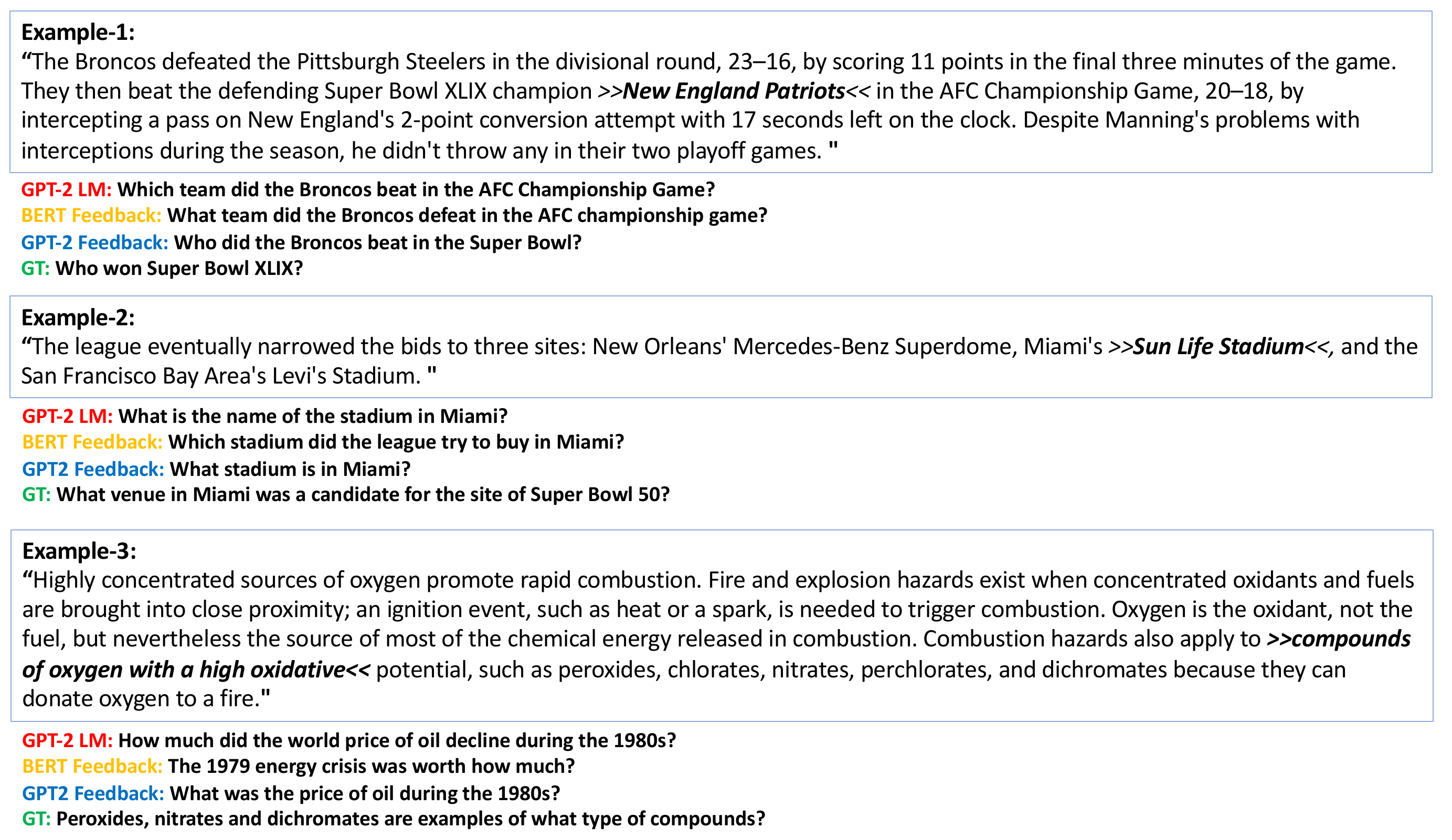}
\captionof{figure}{Some qualitative examples of the questions generated by the proposed method.  Each box contains the context. Within the context the answer tag is delimited by ``$>>$'' and ``$<<$''. Colored text denotes question generated with different language models. ``GPT-2 LM'' corresponds to the GPT-2 LM fine-tuned for question generation without optimization. ``BERT-Feedback'' corresponds to the approach using BERT as QA feedback module during training. ``GPT-2 Feedback'' corresponds to the
approach employing GPT-2 as feedback mechanism. ``GT'' stands for groundtruth. Diversity in answers generated by the proposed approach the semantic richness of the question generation.}
\label{fig:qual}
\end{figure*}

In order to gauge the performance of automatic QG systems, it is very important to have a good metrics at hand. Scores such as BLEU and ROGUE are only of limited use therein, as they mainly measure the lexical similarity between the generated question and the ground truth~\cite{DBLP:journals/corr/TangDQZ17}. That can be attributed to the fact that often these metrics originate a specific application domain, e.g. BLEU for translation. As a result, they are of limited us for other NLP applications. Specifically in the context of question generation they tend to be inadequate as they are unable to capture whether a generated question really looks like a semantic valid question or not. A desirable evaluation system should also have the ability to judge whether the generated question could be answered by input sentence, even if the generated question use totally different words to express the meaning. As an example, taking the first sample from Tab.~\ref{fig:qual}, \emph{``What team did the broncos defeat in the AFC championship game?''} is a perfectly reasonable question given the answer \emph{``New England Patriots''} and the specific context. However, it scores very low in terms of BLEU against the groundtruth \emph{``Who won Super Bowl XLIX?''}, highlighting the deficiency of these scores for the task of assessing QG. 
Motivated by this, we introduced to train a QA system on generated questions by QG system, and utilize the performance of the former model as a surrogate measure for the latter one. For this purpose, we train a BERT QA model on generated questions, as well as combination of generated questions and groundtruth questions. The underlying idea is that if the model is able to generate questions with high diversity using words with low lexical similarity to the groundtruth, the QA system is also improved. Incorporating QA, entails multiple aspects: semantic information as well as answerability - therefore providing complementary cues. Specifically, the QA network becomes more robust as it learns to generalize better. Specifically, in terms of BERT, this implies a widening of the semantic spectrum of the context-specific embedding driving QA model. For all the evaluations of the surrogate metric, BERT was trained for 2 epochs.

As can be seen in Tab.~\ref{tab:results-squad-generative}, the performance in question answering using generated questions using BERT in the feedback loop almost reaches groundtruth benchmark performance. This is followed by using GPT-2 in the feedback loop and by a large margin GPT-2 language model directly for question generation.
At the same time, the strong performance suggests that there is sufficient diversity in the questions generated as well as that the questions are semantically correct. This fact is further highlighted by simultaneously obtained low scores for BLEU and ROGUE metrics in Tab.~\ref{tab:results-quantitative-metrics}. Scores of similar low BLEU magnitude, however, on a slightly different data split are reported in for their related approach~\cite{DBLP:journals/corr/TangDQZ17}.

To better analyze the power of our question generation in terms of semantic diversity, we provide additional groundtruth data for learning the QA model. The rationale behind that was to check if the augmented generated question can also be beneficial in presence of groundtruth data or not. For that, a BERT QA model is trained on entire training set of SQuAD, but half the training data is fully supervised (i.e., contain question / answer pairs), but the other half is not annotated with the questions, and we use our proposed method to generate question for the latter part. Finally, performance is evaluated on the development set of SQuAD. As can be seen in Tab.~\ref{tab:results-squad_full}, the performance in this setup using the questions generated by our method almost reaches to the fully supervised baseline where has been trained on whole training set in a fully supervised manner. The small margin between our QA method, and fully supervised baseline suggests the applicability of our proposed framework for the low data regime scenarios. The large margin between our proposed method and the initialization model clearly shows the semantic diversity of the generated question compared to the groundtruth questions.

\subsection{Semi-Supervised Learning}
In this experiment we study the performance in a semi-supervised setup at different labeling rates. For the sake of transparency, we conducted experiments following the protocol of~\cite{yang-etal-2017-semi} with the associated customized SQuAD split. Specifically, we evaluate the performance at labeling rate of 10\%, 20\%, 50\% and 90\% of the 50k unlabeled data corpus with remaining 10\% used for testing. The results are reported in Table~\ref{tab:results-squad_split-1}. As it can be seen in the table, the proposed approach outperforms the state-of-the-art semi-supervised QA approach of~\cite{yang-etal-2017-semi} by a large margin at any labeling rate.  
The larger the labeling rate is, the larger the margin.  However, it should be noted a part of the margin can be attributed to the approach of~\cite{yang-etal-2017-semi} employs some heuristic to extract answer candidates from which questions are generated, rather than employing groundtruth answer spans. 
The results further suggest that the approach at a labeling rate of around 50\% already starts to saturate. That is, the gain in accuracy beyond that labeling rate drops significantly, indicating the high quality and diversity of the generated questions.
Generally, using the proposed approach in a semi-supervised setting seems feasible given that even at very low labeling rates such as 10\%, the proposed approach reaches good accuracy not too far off from the upper bound.

\subsection{Ablation Study of QA component}
\label{sec:ablation}
The proposed approach employs BERT as QA component for co-training the QG component. In this section we analyze the effect of the type of QA system in collaboration with QG system. Specifically, we replace BERT with a variant of GPT-2 that emulates BERT. In order achieve question answering emulating span regression in BERT-like fashion, the GPT-2 architecture has to be altered. To this end, a ``regression-head'' has to be added at the top of GPT-2 stack.  As a result, one obtains a trainable QA head layer that facilitates association of log-likelihoods for each context token, indicating the probability of being a span delimiter. It should be noted that in the original GPT-2, answers were generated using the language model instead. Table\ref{tab:results-squad-generative} shows the results of using the proposed approach that employs BERT as QA module during training vs. using the modified GPT-2 QA variant. All models were fine-tuned for 2 epochs. In both cases (a separate) BERT QA module was used as surrogate evaluation metric as introduced in the previous subsection. The results suggest that using BERT leads to a much better performance than GPT-2 QA. This can be attributed that GPT-2 features a language model, with its backbone not optimized for multiple downstream tasks. Therefore it can provide only limited and unreliable feedback in terms of diversity of the questions generated. In contrast to that BERT, with its context-specific embeddings allows for robust and reliable QA.



\section{Conclusion}
In this paper, a simple yet effective approach for question generation is presented. For that, we leverage question generation by tying together GPT-2 and BERT in end-to-end trainable fashion facilitating semi-supervised learning. The generated questions are of high quality, showing high semantic similarity w.r.t. groundtruth data. Furthermore, conducted experiments suggest that the proposed approach allows for generation of question significantly reducing the burden of full annotation. BERT as QA module in the feedback loop model shows best performance, which may be attributed to the bi-directional context specific embeddings leveraging a powerful feedback mechanism.  Additionally, as the BLEU and similar scores are weak metric for assessing generative power, we proposed to use BERT QA as a surrogate measure for assessing question generation quality. Future work will entail on further improving question generation as well as reducing the requirements of answer annotations. Completely eliminating the answer annotations could pave the way towards fully unsupervised question generation.

\bibliography{aaai20}
\bibliographystyle{aclnatbib}
\end{document}